\documentclass{article}

\usepackage{tcolorbox}
\definecolor{darkblue}{rgb}{0,0.25,0.75}
\usepackage[hidelinks]{hyperref}
\hypersetup{
  colorlinks=true,
  linkcolor=darkblue,
  citecolor=darkblue,
  urlcolor=darkblue
}

\usepackage{amsmath,amsfonts,amsthm,amssymb,bm}

\def\eqref#1{equation~\ref{#1}}

\def\1{\bm{1}}

\def\vx{{\mathbf{x}}}

\DeclareMathAlphabet{\mathsfit}{\encodingdefault}{\sfdefault}{m}{sl}
\SetMathAlphabet{\mathsfit}{bold}{\encodingdefault}{\sfdefault}{bx}{n}

\newcommand{\R}{\mathbb{R}} %

\theoremstyle{definition}

\theoremstyle{definition}

\usepackage{style}

\usepackage{wrapfig}

\usepackage[preprint]{neurips_2026}

\usepackage[utf8]{inputenc} %
\usepackage[T1]{fontenc}    %
\usepackage{hyperref}       %
\usepackage{url}            %
\usepackage{booktabs}       %
\usepackage{amsfonts}       %
\usepackage{nicefrac}       %
\usepackage{microtype}      %
\usepackage{xcolor}         %
\usepackage{graphicx}

\definecolor{orange}{rgb}{1,0.5,0}
\definecolor{mdred}{rgb}{0.7,0,0}
\definecolor{mdgreen}{rgb}{0.05,0.6,0.05}
\definecolor{mdblue}{rgb}{0,0,0.7}
\definecolor{dkblue}{rgb}{0,0,0.5}
\definecolor{dkgreen}{rgb}{0,0.5,0}
\definecolor{dkgray}{rgb}{0.3,0.3,0.3}
\definecolor{slate}{rgb}{0.25,0.25,0.4}
\definecolor{gray}{rgb}{0.5,0.5,0.5}
\definecolor{ltgray}{rgb}{0.7,0.7,0.7}
\definecolor{purple}{rgb}{0.7,0,1.0}
\definecolor{lavender}{rgb}{0.65,0.55,1.0}

\newcommand{\modelname}{\textgreater~\textless former\xspace}

\title{Variable-Width Transformers}

\author{Zhaofeng Wu$^\text{\Cancer}$ \quad Oliver Sieberling$^\text{\Cancer}$ \quad Shawn Tan$^\text{\Sagittarius}$ \\ \textbf{Rameswar Panda}$\text{\Sagittarius}$ \quad \textbf{Yury Polyanskiy}$^\text{\Cancer}$ \quad \textbf{Yoon Kim}$^\text{\Cancer}$ \\
$^\text{\Cancer}$MIT \quad
$^\text{\Sagittarius}$MIT-IBM Watson AI Lab \\
\texttt{zfw@csail.mit.edu}
}

\begin{document}

\maketitle

\begin{abstract}
Scaling model size, specifically depth and width, has driven significant progress in transformer-based language models. However, most architectures maintain a constant width across all layers, allocating a fixed parameter and computation budget evenly despite different layers potentially playing distinct computational roles. In this work, we empirically investigate nonuniform capacity allocation across network depth by proposing a $\times$-shaped \modelname architecture. This design maintains wider early and late layers while narrowing the middle layers, utilizing a parameter-free residual resizing mechanism. Across decoder-only language models ranging from 200M to 2B parameters (dense) and 3B parameters (MoE), our \modelname consistently outperforms parameter-matched uniform baselines on language modeling loss. By reducing the average layer width, this architecture also requires fewer overall FLOPs (22\% reduction under fitted loss-matched scaling curves) and smaller KV cache memory and I/O cost (15\% reduction). In analysis, we show that this bottleneck structure results in qualitatively different representations in residual streams. Overall, our results demonstrate that nonuniform width allocation can result in more resource-optimal scaling of language models.
\end{abstract}

\blfootnote{We release our code at \url{https://github.com/ZhaofengWu/variable-width-transformers}.}

\begin{figure}[h]
    \centering
    \includegraphics[width=0.5\linewidth]{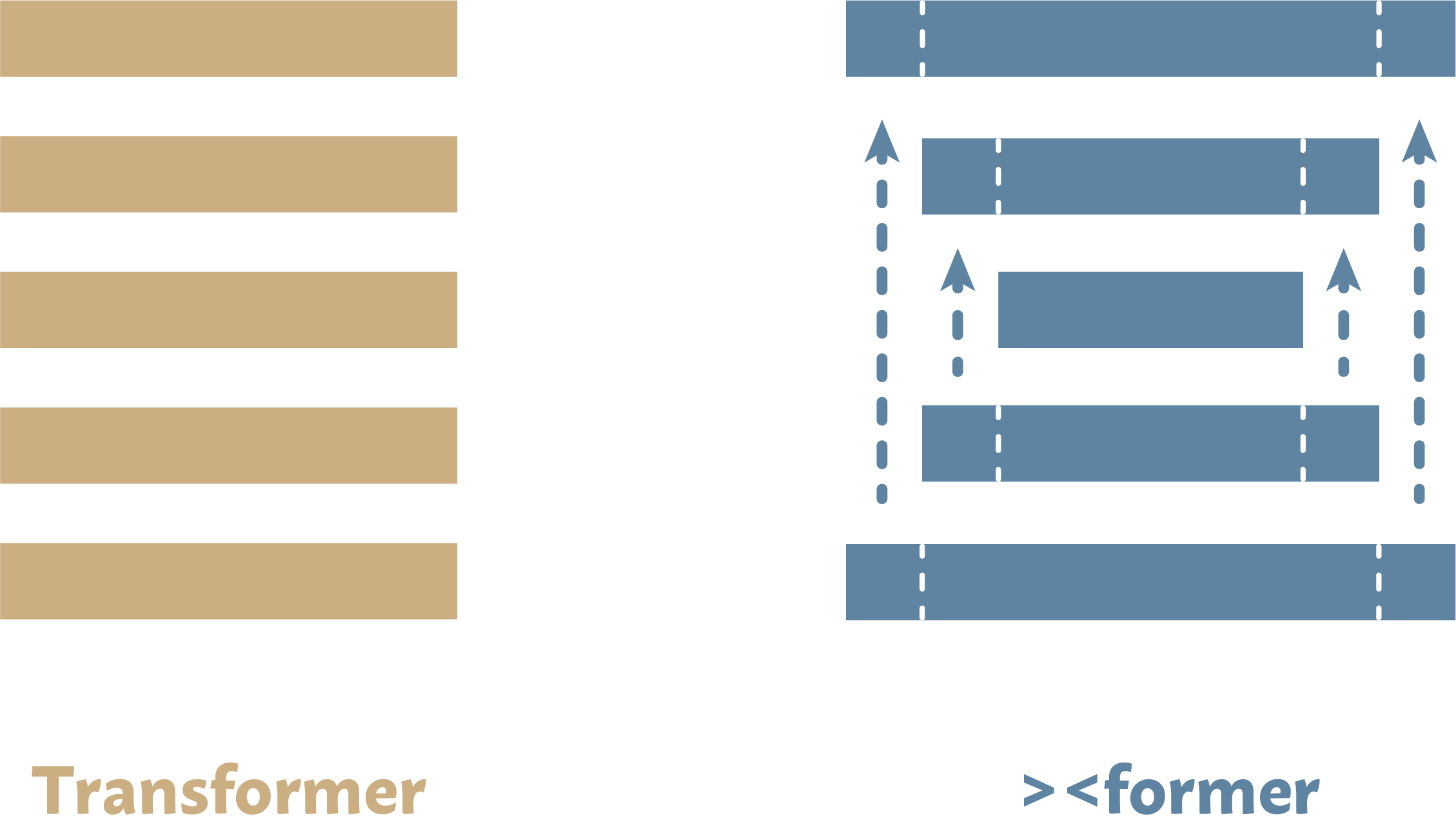}
    \caption{We propose \modelname, where different layers have different widths. We specifically employ a $\times$-shaped architecture, where inactive dimensions are copied upward in the residual stream. We find that this improves performance and saves on training FLOPs, KV cache memory, and I/O cost.
    }
    \label{fig:overview}
\end{figure}

\section{Introduction}

Scaling has been a critical driver of progress in modern AI. One major axis of scaling is model size. In the context of transformers, model size is a function of the dimension of the transformer block (a model's ``width'')\footnote{The intermediate MLP dimension is typically a fixed multiple of the model dimension (though see \citealp{ikeda2025layerwise}).} and the number of transformer blocks (a model's ``depth'').
Consequently, much prior work has investigated how to optimally scale model size through scaling transformer width and depth. While early research suggested that a model's shape (i.e., the ratio of width to depth) mattered less than the total parameter count for performance \citep{kaplan2020scalinglawsneurallanguage}, subsequent work found that model shape can lead to nontrivial differences \citep{levine2020limits,tay2021scale,petty2024impact} and should be taken into account when fitting scaling laws \citep{mcleish2025gemstones}. 

Yet, even as the optimal global shape of transformers is debated, these studies generally preserve a less-examined assumption: a model's width is constant across depth. That is, once a hidden dimension size is chosen, every transformer block receives approximately the same computation/parameter budget. This uniform-width design is convenient, but not obviously optimal. Different layers may play different roles during computation, and a fixed total parameter or FLOP budget need not be allocated evenly across depth. This motivates a general question: under a fixed depth and parameter budget, should all layers have the same width, or should capacity be distributed nonuniformly?

We study this question empirically by training decoder-only transformer language models (LMs) with nonuniform allocation of parameters and compute across depth. Concretely, for a given parameter and depth constraint, we vary the model shape across several settings: growing ($\vee$-shaped), narrowing ($\wedge$-shaped), growing then narrowing ($\Diamond$-shaped), and narrowing then growing ($\times$-shaped). Across these settings, we find that $\times$-shaped models (wide in early and late layers but narrower in the middle) outperform parameter-matched constant-width transformers. We call them \modelname{s}. This differs from prior work on layerwise allocation of only FFN intermediate dimensions, which found benefits from allocating more computation to middle layers \citep{ikeda2025layerwise}; our results instead suggest that reallocating the full block width leads to a different optimal profile. 

A key implementation detail is how variable-width layers interact with the residual stream. Na\"{i}vely changing the residual dimension between layers introduces projection bottlenecks and changes the skip path. We instead keep a fixed global residual dimension and allow each block to read from and write to a layer-specific slice of the residual stream. Coordinates not used by a given block bypass that block and are projected upstream via copying. We find that this fixed-residual construction is important for realizing the gains from nonuniform width profiles.

Nonuniform width allocation also has efficiency benefits: it requires fewer training and inference FLOPs than constant-width transformers, while also reducing the KV cache memory and I/O cost for moving activations. Because a layer's parameter count scales quadratically with its width, while attention FLOPs and KV cache size scale linearly, matching the parameter count of a uniform baseline results in a reduction in average layer width (and hence the KV cache). Across models with 200M--2B parameters, \modelname{s} achieve approximately a 3\% relative improvement in perplexity over parameter-matched constant-width baselines while reducing KV-cache size by about 10\% and FLOPs by about 3\%. These benefits also extend to mixture-of-experts transformers. We further analyze how the optimal bottleneck width and bottleneck location depend on the model budget, providing empirical guidance for scaling nonuniform-width transformers. Finally, we also perform analyses to understand the benefit of \modelname, showing that it employs a different representation strategy than the constant-width baseline and mitigates mid-layer representation collapse.

\section{Variable-Width Transformers} \label{sec:method}

A standard transformer contains a series of $L$ layers. In each layer $\ell\in[1,L]$, a transformer block $\mathcal{B}^\ell: \R^d \to \R^d$ transforms the input from the previous layer $\vx^{\ell-1}$ by $\vx^\ell = \mathcal{B}^\ell(\vx^{\ell-1}) + \vx^{\ell-1}$. $d$ is the model dimension.
We define $\vx^0$ as the input embeddings.

In this work, we question why $d$ must be held constant. Much past work has shown that different layers of a transformer LM perform distinct functions, which naturally may require different amounts of capacity~\citepia{tenney-etal-2019-bert,meng2022locating,10.1016/j.csl.2022.101429}. This motivates each layer $\ell$ having a different dimension $d_\ell$.

One practical challenge, however, is that this requires resizing between layers: $\vx^\ell = \mathcal{B}^\ell(f^{\ell}(\vx^{\ell-1})) + f^\ell(\vx^{\ell-1})$ where $f^\ell: \R^{d_{\ell-1}}\to\R^{d_{\ell}}$ resizes the hidden state.
We consider a parameter-free approach.
When shrinking dimensions, i.e., $d_\ell < d_{\ell - 1}$, we simply truncate the extra dimensions, i.e., $f^\ell(\vx) = \vx[:d_\ell]$.
When expanding dimensions, i.e., $d_\ell > d_{\ell - 1}$, we restore each previously truncated dimension from the most recent layer that actively processed it. Formally, for each coordinate index $i \in \{1, \dots, d_\ell\}$, the $i$-th element of the resized hidden state $f^\ell(\vx^{\ell-1}) \in \mathbb{R}^{d_\ell}$ is constructed as:
\begin{equation}
    [f^\ell(\vx^{\ell-1})]_i = [\vx^{\ell'}]_i \quad \text{where} \quad \ell' = \max \{ \tilde{\ell} < \ell \mid d_{\tilde{\ell}} \ge i \}
\end{equation}
If no such prior layer exists (i.e., if the required dimension exceeds the maximum width of all preceding layers), the coordinate is padded with $0$.
See \S\ref{sec:moveup-ablation} for ablations that show that these methods outperform alternatives such as training a projection layer or always padding with 0s.

Using this expansion method, we can mathematically conceptualize our variable-width model as a uniform-width model except (1) each layer only reads from/writes to a subset of residual stream dimensions, and (2) it has a larger residual stream width (equal to the width of the widest layer).

We investigate different shapes with two additional parameters, $\ell^*$, the layer index of a bottleneck layer, and $d_{\ell^*}$, its dimension.
We parameterize the rest of the layer widths geometrically:\footnote{A geometric layer width schedule outperforms an arithmetic one in preliminary experiments.} $d_\ell = \alpha^- d_{\ell-1}$ with change rate $\alpha^{-}$ for $\ell\le\ell^*$ (the early layers) and $d_\ell = \alpha^+ d_{\ell-1}$ with change rate $\alpha^{+}$ for $\ell>\ell^*$ (the late layers).
With different settings of $\alpha^+$, $\alpha^-$, and $\ell^*$, we recover different kinds of shapes:
when $\alpha^-<1$ and $\alpha^+>1$, we obtain a $\times$-shaped model;
when $\alpha^->1$ and $\alpha^+<1$, we obtain a $\Diamond$-shaped model;
when $\ell^*$ is 1 or $L$, we obtain a $\vee$ or $\wedge$-shaped model.
We keep the size of the input and output embeddings unchanged: the QKV projections of the first layer and the MLP down-projection of the last layer adjust for these size mismatches.\footnote{The residual connection in the final layer MLP is truncated accordingly.}

A compelling property of our variable-width architecture is that, when matched in parameter count to a constant-width baseline, it requires strictly fewer overall FLOPs and has a strictly lower average layer width (and thus, lower KV cache size and lower I/O cost for moving activations). Concretely, 
the parameter count of a transformer layer is dominated by the linear projection matrices (QKV, output, and MLP), which scale quadratically with the hidden dimension: $P_\ell \approx K d_\ell^2$, where $K$ is a constant depending on the number of projection matrices and the MLP expansion factor.
Therefore, if we match the parameter count of a variable-width model to a baseline of constant width $d$, we effectively equate the sum of the squared dimensions:
$$ K \sum_{\ell=1}^L d_\ell^2 = K L d^2 \implies \frac{1}{L} \sum_{\ell=1}^L d_\ell^2 = d^2. $$
Because
the square of the mean is upper-bounded by the mean of the squares, and because variable-width ensures the widths $d_\ell$ are not constant, the average layer size is strictly smaller:
$$ \left( \frac{1}{L} \sum_{\ell=1}^L d_\ell \right)^2 < \frac{1}{L} \sum_{\ell=1}^L d_\ell^2 = d^2 \implies \frac{1}{L} \sum_{\ell=1}^L d_\ell < d. $$
For FLOPs, first of all, the number of FLOPs per token in a linear projection is strictly proportional to the number of weights, and so when parameter-matched, the total dense FLOPs remain identical to the baseline.
For attention dot-products, their FLOPs scale linearly with the hidden dimension: $\text{FLOPs}_\ell \propto N^2 d_\ell$, where $N$ is the sequence length.
Therefore, the total attention compute $\sum_{\ell=1}^L N^2 d_\ell=N^2 \sum_{\ell=1}^L d_\ell$ is consequently strictly lower than the baseline $N^2 L d$.\footnote{In practice, adjusting for the sizes of the input and output embeddings (see above; formalized in \S\ref{app:param-match}) introduces a minor parameter correction. Nevertheless, this $O(1)$ boundary effect is heavily dwarfed by the $O(L)$ bulk layer parameters, preserving these theoretical results under meaningful width schedules.}

In summary, there are 4 parameters for a variable-width transformer, $\ell^*$, $d_{\ell^*}$, $\alpha^+$, and $\alpha^-$. We set two constraints: $d_1=d_L$ (for $\Diamond$ and $\times$ shapes) and that the parameter count matches a constant-width baseline.\footnote{As shown in the proof, we cannot match the parameter count and the FLOP count at the same time.} So for our experiments, we consider $\ell^*$ and $d_{\ell^*}$ as two hyperparameters, and automatically solve for all layer widths.\footnote{After solving the layer widths, we post-hoc round each width to the nearest 32, which is the number of attention heads we use (16) times 2 (for RoPE~\citep{10.1016/j.neucom.2023.127063}).} See \S\ref{app:param-match} for our derivation.

\section{\modelname}

\begin{table}[t]
\centering
\caption{Pre-training hyperparameters.}
\label{tab:training-hparams}
\small
\begin{tabular}{lc@{\hspace{3mm}}c@{\hspace{3mm}}c@{\hspace{3mm}}c@{\hspace{3mm}}c@{\hspace{3mm}}c}
\toprule
Parameters & Layers ($L$) & Hidden ($d$) & Batch Size & Tokens & Experts (Tot/Act) & MLP Interm. Size \\
\midrule
\multicolumn{7}{c}{\textit{Dense Models}} \\
\midrule
200M & 16 & \phantom{0}640 & \phantom{0}512 & \phantom{0}10B & -- & -- \\
500M & 24 & \phantom{0}960 & 1024 & \phantom{0}25B & -- & -- \\
1B   & 32 & 1280         & 2048 & \phantom{0}50B & -- & -- \\
2B   & 40 & 1600         & 4096 & 100B & -- & -- \\
\midrule
\multicolumn{7}{c}{\textit{Mixture of Experts}} \\
\midrule
3B (1B active) & 40 & 1600 & 4096 & 100B & 22 / 3 & 512 \\
\bottomrule
\end{tabular}
\end{table}

In this section, we discuss training variable-width transformers. After introducing our training setup, we first establish that a $\times$-shaped model works best, and then identify a parameterization of the bottleneck layer index $\ell^*$ and dimension $d_{\ell^*}$ that works well across model sizes.
Using this recipe, we pre-train \modelname{s} and constant-width baselines across sizes and find that \modelname{s} consistently achieve better loss and downstream task performance with a smaller pre-training FLOPs footprint and KV cache size.

\subsection{Training Setup} \label{sec:setup}

We pre-train four model sizes---200M, 500M, 1B, and 2B---with different numbers of layers and hidden sizes (Table \ref{tab:training-hparams}). For each, we pre-train constant-width transformers and variable-width models.
We also consider a Mixture-of-Experts (MoE) model with 3B total/1B active parameters. For parameter-matching the variable-width MoE model with the baseline, we match the number of total parameters---this results in the variable-width model having 3\% fewer active parameters, but we show in \S\ref{sec:results} that it still outperforms the constant-width baseline despite this.

For pre-training data, we train on DCLM~\citep{li2024datacomplm}. For each model size, we train models to 2.5$\times$ Chinchilla-optimal~\citep{10.5555/3600270.3602446}, i.e., the number of trained tokens is equal to 50 times the parameter count, e.g., 100B tokens for the 2B model. We train on length-4096 sequences, with model-size-dependent batch sizes (Table \ref{tab:training-hparams}). Inputs are tokenized with OpenAI's \texttt{cl100k\_base}.\footnote{\url{https://github.com/openai/tiktoken}}

All models are trained with maximal update parametrization ($\mu$P; \citealp{yang2024tensor}). We use $\mu$P-aware initialization and optimizer parameter groups, with the same AdamW hyperparameters across scales: learning rate $10^{-2}$, $\beta=(0.9,0.95)$, weight decay $0.1$, and $\epsilon=10^{-10}$.
We use a power learning-rate decay schedule. The learning rate is linearly warmed up for approximately the first 8\% of training steps and then decayed for the remaining steps. All models are trained in bfloat16 precision.
Following common practice, we omit bias terms from all linear projections, including attention, MLP, and output projections~\citepia{chowdhery2022palmscalinglanguagemodeling,groeneveld-etal-2024-olmo}. We use the SwiGLU activation~\citep{shazeer2020gluvariantsimprovetransformer}.

We measure the training loss of each model,\footnote{Because we do not repeat any data, this is in expectation the same as held-out loss.} averaged over the final 1,000 steps (with a 10-step increment) for smoothing.
We also report pre-training FLOPs in PFLOP/s-days, the number of days required assuming 1 PFLOP per second~\citep{kimiteam2026attentionresiduals}.
Finally, we compare the average layer size, which is proportional to the KV cache size during inference.

\subsection{The $\times$ Shape Works Best}

\begin{figure}[t!]
    \centering
    \includegraphics[width=0.6\linewidth]{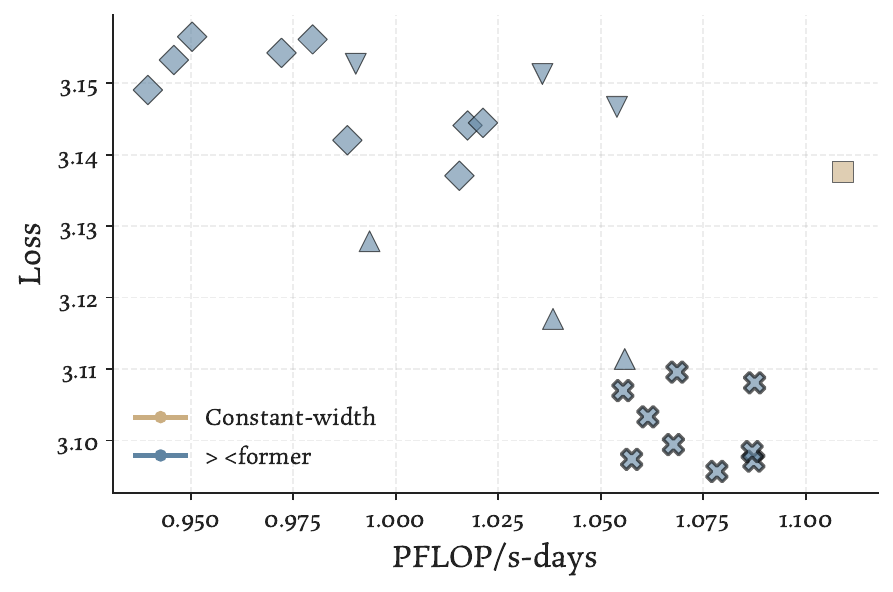}
    \caption{Comparing variable-width transformers with different shapes, each sweeping over multiple hyperparameter choices. \textbf{The $\times$-shaped model performs the best.}}
    \label{fig:shapes}
\end{figure}

We first explore different shapes on the 500M-parameter scale. Beyond a regular constant-width transformer, we experiment with a $\Diamond$ shape, a $\times$ shape, a $\vee$ shape, and a $\wedge$ shape. Due to the optimal hyperparameter ($\ell^*$ and $d_{\ell^*}$, \S\ref{sec:method}) potentially differing per shape, we experiment with 3 choices for each hyperparameter per shape. This amounts to 9 runs for the $\times$ shape and the $\Diamond$ shape, and 3 runs for the $\vee$ shape and the $\wedge$ shape, for which $\ell^*$ is irrelevant. In Figure~\ref{fig:shapes}, we plot their loss and the pre-training FLOPs. We see that the $\times$ shape consistently performs the best.\footnote{Anecdotally, our initial intuition was to pursue a $\Diamond$-shaped model, increasing the computation in middle layers which are often associated with semantic computations~\citep{tenney-etal-2019-bert}. We nevertheless proceed with $\times$-shaped models due to these empirical results.}

\subsection{Finding a Specific Width Schedule}

\begin{figure}[t!]
    \centering
    \includegraphics[width=\linewidth]{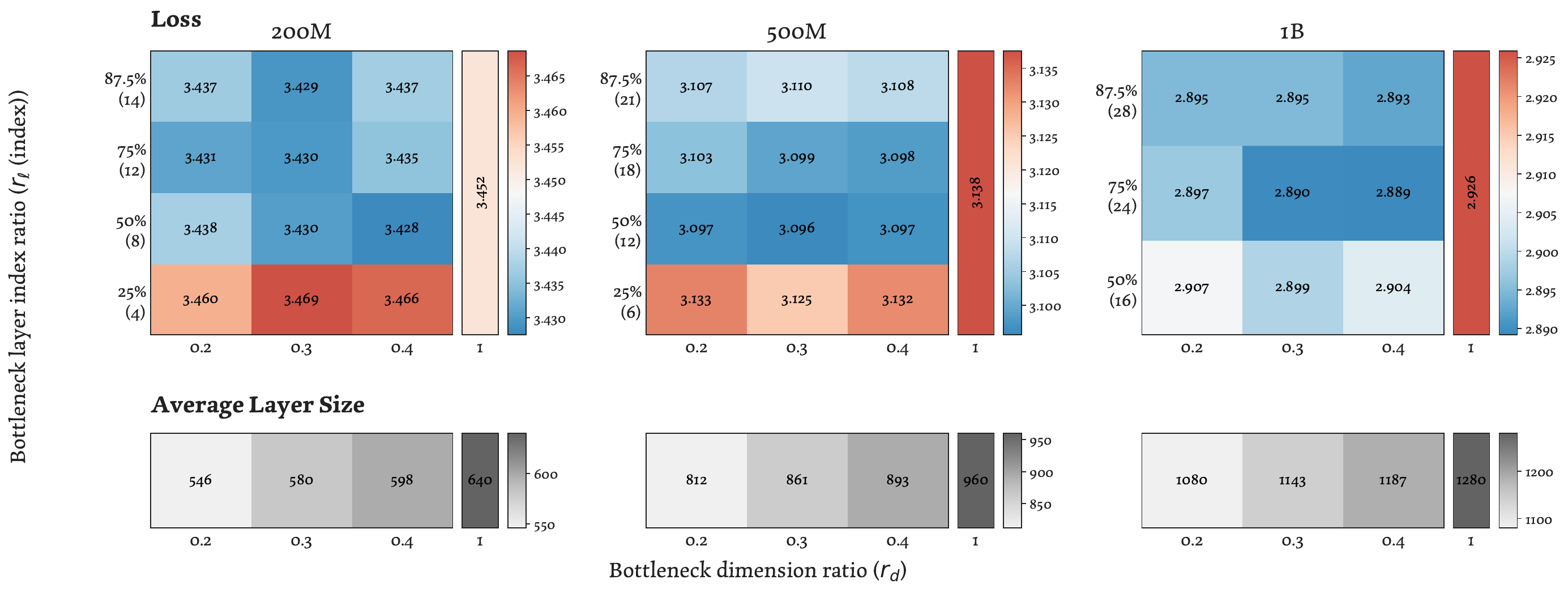}
    \caption{The effect of the bottleneck layer index and dimension on language modeling loss, parameterized as a ratio to the total number of layers and the base dimension. $\ell^*=r_\ell L$ and $d_{\ell^*}=r_d d$. 
    We also show the baseline performance, indicated using $r_d=1$, and the resulting average layer size.
    \textbf{This parameterization yields a relatively consistent model performance pattern across model sizes.}}
    \label{fig:hparams}
\end{figure}

As mentioned in \S\ref{sec:method}, we need to choose a bottleneck layer index $\ell^*$ and the bottleneck dimension $d_{l^*}$. Ideally, we want to find a recipe that works well across model sizes so that we do not have to search for them individually at each model size. Therefore, we parameterize these two hyperparameters as ratios to the total number of layers $L$ and the hidden size $d$: $\ell^*=r_\ell L$ and $d_{l^*}=r_d d$. We sweep over different values of $r_\ell$ and $r_d$ at small model sizes: 200M, 500M, and 1B. Figure~\ref{fig:hparams} shows the results. While it is not the case that a single $(r_\ell, r_d)$ pair is consistently the best, the fact that such a ratio-based parameterization leads to \emph{roughly} similar trends across model sizes is interesting. By default, based on this sweep, we use $\ell^*=0.75L$ and $d_{l^*}=0.3d$ going forward.

\subsection{\modelname Outperforms Constant-Width Transformer} \label{sec:results}

\begin{table}[t!]
\centering
\small
\caption{The performance, pre-training FLOPs, and average layer size of \modelname{s} vs. constant-width transformers. \textbf{\modelname consistently achieves lower loss with lower pre-training FLOPs and average layer size.}}
\begin{tabular}{lllll}
\toprule
Size & Model & Loss & PFLOP/s-days & Avg layer size \\
\midrule
\multirow{2}{*}{200M} & Transformer & 3.452 & 0.18 & 640 \\
 & \modelname & \textbf{3.430} & \textbf{0.17 ($-$3.2\%)} & \textbf{576 ($-$10.0\%)} \\
\midrule
\multirow{2}{*}{500M} & Transformer & 3.138 & 1.11 & 960 \\
 & \modelname & \textbf{3.099} & \textbf{1.07 ($-$3.7\%)} & \textbf{855 ($-$11.0\%)} \\
\midrule
\multirow{2}{*}{1B} & Transformer & 2.926 & 4.52 & 1280 \\
 & \modelname & \textbf{2.890} & \textbf{4.41 ($-$2.6\%)} & \textbf{1145 ($-$10.5\%)} \\
\midrule
\multirow{2}{*}{2B} & Transformer & 2.751 & 16.92 & 1600 \\
 & \modelname & \textbf{2.726} & \textbf{16.49 ($-$2.5\%)} & \textbf{1426 ($-$10.9\%)} \\
\midrule
 \multirow{2}{*}{3B/1B MoE} & Transformer & 2.726 & 10.13 & 1600 \\
 & \modelname & \textbf{2.710} & \textbf{9.66 ($-$4.6\%)} & \textbf{1426 ($-$10.9\%)} \\
\bottomrule
\end{tabular}
\label{tab:results}
\end{table}

\begin{figure}[t!]
    \centering
    \includegraphics[width=\linewidth]{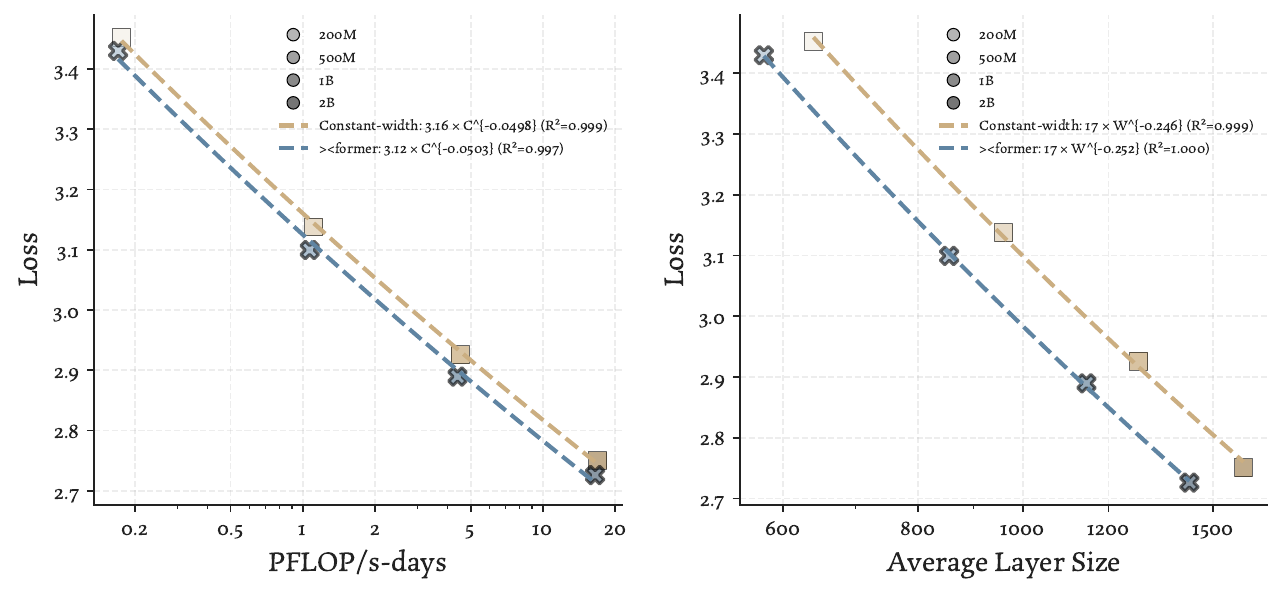}
    \caption{Language modeling loss vs. pre-training FLOPs (left) and average layer size (right). \textbf{\modelname produces lower loss at smaller FLOP and average layer size costs.}}
    \label{fig:main-results}
\end{figure}

\begin{table*}[t]
\centering
\small
\resizebox{\textwidth}{!}{
\begin{tabular}{lcccccccccccccc}
\toprule
& \multicolumn{12}{c}{Accuracy ($\uparrow$)}
& \multicolumn{2}{c}{Perplexity ($\downarrow$)} \\
\cmidrule(lr){2-13}\cmidrule(lr){14-15}
Model
& ARC-C
& ARC-E
& BoolQ
& COPA
& HellaSwag
& LAMBADA
& OBQA
& PIQA
& RACE
& SciQ
& WinoGrande
& Avg.
& LAMBADA
& WikiText \\
\midrule
2B constant-width
& 33.0 & 59.5 & 59.4 & 76.0 & 55.9
& 55.4 & 33.8 & 73.3 & 34.3 & 79.5 & 57.0
& 56.1
& 8.18 & 16.96 \\
2B \modelname
& 34.4 & \textbf{63.3} & 60.9 & 73.0 & \textbf{57.9}
& 56.1 & 33.6 & 74.4 & 33.4 & 82.0 & \textbf{60.2}
& \textbf{57.2}
& \textbf{7.43} & \textbf{16.32} \\

\midrule

MoE baseline
& 33.7 & 62.2 & \textbf{63.0} & 77.0 & 57.3 & 56.1
& 37.4 & 74.8 & 34.6 & \textbf{83.9} & 55.6 & 57.8
& 7.78 & 16.36 \\
MoE \modelname
& 33.2 & 61.0 & 59.5 & 80.0 & \textbf{58.7} & 56.2
& 37.8 & 75.2 & 34.3 & 80.2 & \textbf{60.1} & 57.8
& \textbf{7.45} & \textbf{15.98} \\
\bottomrule
\end{tabular}
}
\caption{Model performance on standard LM evaluation datasets. For accuracy metrics, bold indicates the higher value when significant at $p < 0.05$ under a one-sided test; for perplexity, bold indicates the lower value. \textbf{\modelname{s} consistently outperform constant-width transformers on perplexity-based tasks, and the 2B \modelname wins on most natural language understanding tasks.}}
\label{tab:dense-moe-eval}
\end{table*}

Table~\ref{tab:results} shows, at all model sizes that we tested, \modelname outperforms the constant-width transformer, while requiring fewer FLOPs and average layer size (i.e., with a reduction in KV cache size).

In Figure~\ref{fig:main-results} (left), we fit a scaling law curve on loss vs. pre-training FLOPs to \modelname{s} and constant-width transformers~\citep{kaplan2020scalinglawsneurallanguage}, finding a tight fit.
Similarly, in Figure~\ref{fig:main-results} (right), we also find a tight power-law fit on loss vs. the average layer size.
From these scaling law curves, we compute that \modelname can achieve the 2B constant-width transformer's loss (2.751) with 77.8\% FLOPs and 85.1\% average layer width. Furthermore, both scaling law curves show that not only does \modelname have a smaller intercept, but it has a slightly steeper scaling exponent too, suggesting that the gaps might widen at larger sizes.

We also test these models on standard LM downstream evaluation benchmarks using the \texttt{lm-evaluation-harness}~\citep{eval-harness} in the zero-shot setting. This suite covers natural language understanding (NLU) tasks such as common-sense reasoning, reading comprehension, etc., as well as perplexity-based tasks. For multiple-choice tasks, we report normalized accuracy when available, since it corrects for answer-length effects by normalizing choice likelihoods. When it is not provided, we report standard accuracy. We report the dataset statistics and metrics in Table~\ref{tab:downstream-eval-stats} in \S\ref{sec:dataset-stats}. We evaluate the 2B models and the MoE models and show the results in Table~\ref{tab:dense-moe-eval}. \modelname{s} consistently outperform constant-width transformers on perplexity-based tasks. The 2B \modelname also leads on most NLU tasks. The MoE \modelname is mixed on NLU accuracy but improves both perplexity metrics; at these model sizes, we treat perplexity as the more informative metric of LM quality. We again note that \modelname achieves this with fewer FLOPs and memory, and also fewer active parameters for the MoE model (\S\ref{sec:setup}).

\begin{figure}[t!]
    \centering
    \includegraphics[width=\linewidth]{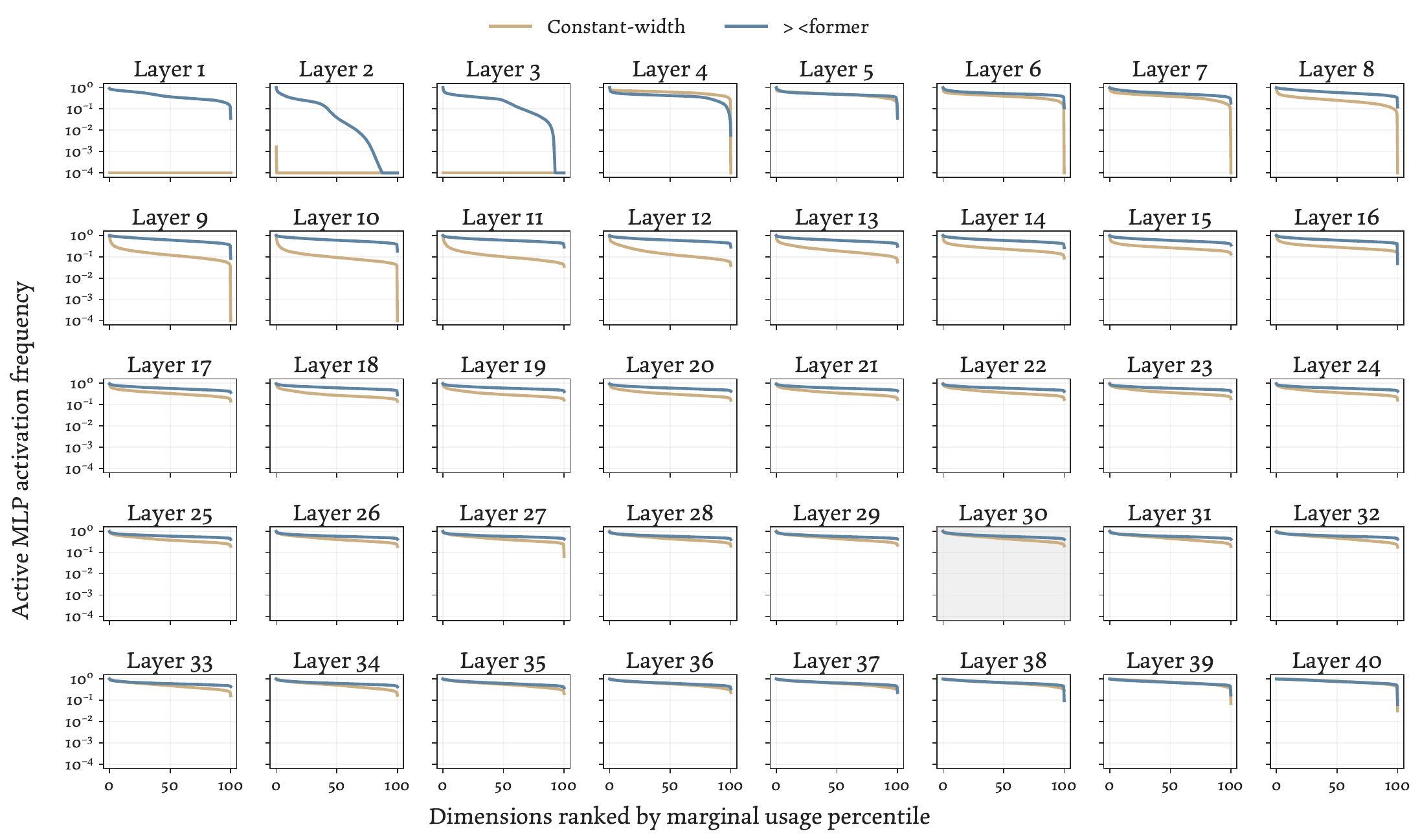}
    \caption{The utilization frequency of MLP activation dimensions in the 2B \modelname vs. the 2B constant-width transformer, visualized separately for each layer. 
    The shaded panel corresponds to the bottleneck layer.
    \textbf{\modelname more evenly utilizes MLP activation dimensions.}}
    \label{fig:activation-utilization}
\end{figure}

\section{Analysis}

Transformers are known to use depths inefficiently~\citep{gromov2025the}, frequently developing ``compression valleys'' where their middle layers collapse in representational capacity and compress computations~\citep{skean2025layer,queipo-de-llano2026attention}.
By inspecting both MLP intermediate activations and the residual stream after each layer, we find that \modelname employs a different representation strategy, where it mitigates the collapse in middle layers and more effectively uses its capacity than a constant-width transformer.

\subsection{\modelname Improves MLP Activation Utilization} \label{sec:analysis-activation}

\begin{figure}[t!]
    \centering
    \begin{minipage}{0.48\textwidth}
        \centering
        \includegraphics[width=\linewidth]{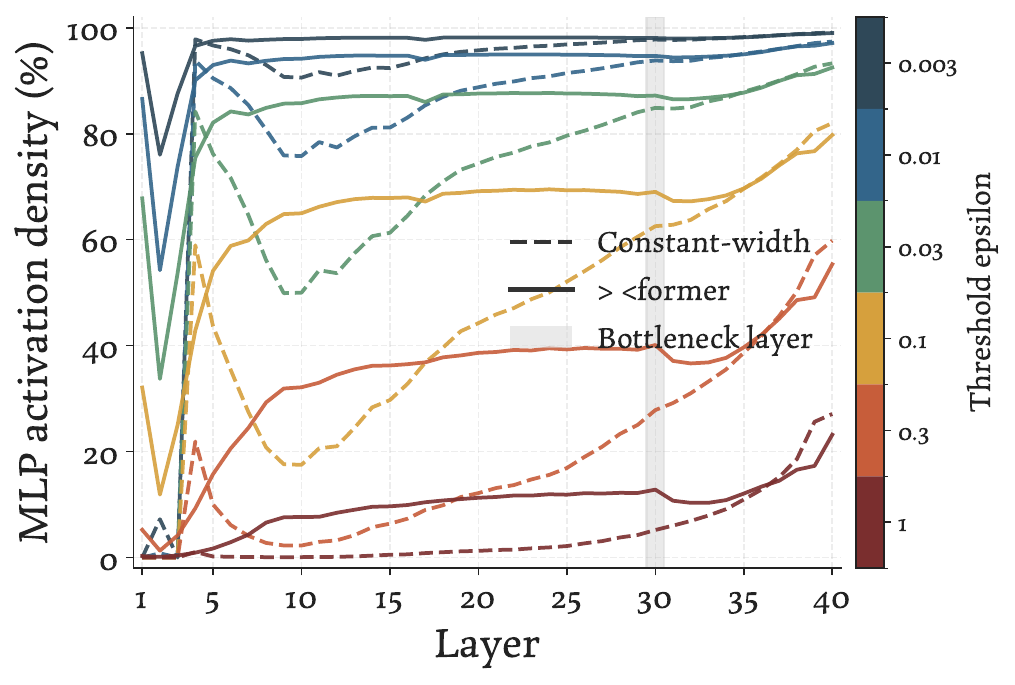}
        \caption{The density of MLP activations in the 2B \modelname vs. the 2B constant-width transformer, across thresholds. \textbf{\modelname consistently more densely activates MLP activation dimensions.}}
        \label{fig:activation-sparsity}
    \end{minipage}\hfill
    \begin{minipage}{0.48\textwidth}
        \centering
        \includegraphics[width=\linewidth]{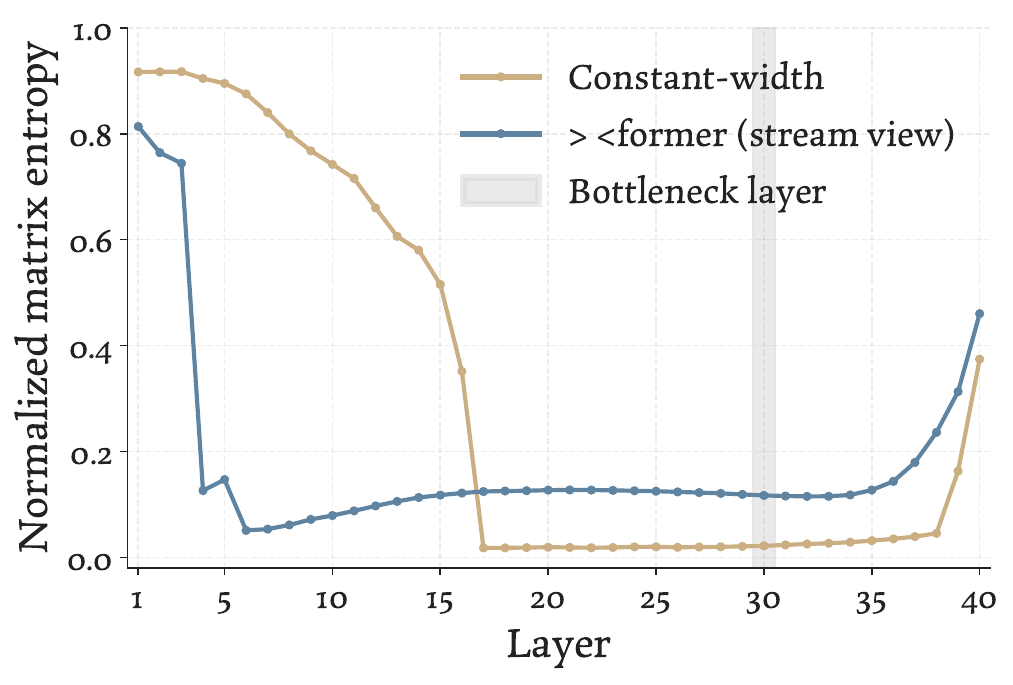}
        \caption{The normalized matrix entropy (\S\ref{sec:analysis-residual}) of layer outputs in the 2B \modelname vs. the 2B constant-width transformer. \textbf{\modelname has a higher matrix entropy in middle-to-final layers, which corresponds to more even usage of the residual dimensions in those layers.}}
        \label{fig:matrix-entropy}
    \end{minipage}
\end{figure}

Intuitively, \modelname enforces an information bottleneck that may encourage the model to more effectively use its representation capacity. We operationalize this by inspecting the utilization of intermediate MLP activations.
Prior interpretability work has viewed a transformer MLP layer as containing key-value memories: the up-projection layer encodes keys for distinct concepts, the intermediate activation represents the MLP input's affinities with the concepts, and the down-projection encodes values for those concepts, taking a linear combination of them weighted by the affinity scores (the activation)~\citep{geva-etal-2021-transformer,geva-etal-2022-transformer}.
We measure the MLP activation density of the 2B \modelname vs. the constant-width transformer on the WikiText-2 validation split with 252,986 tokens~\citep{merity2017pointer}.
Because SwiGLU is continuous, we consider a dimension as active iff its activation magnitude is larger than a certain threshold.
In Figure~\ref{fig:activation-sparsity}, we show that \modelname enforces denser activations within MLPs across thresholds.

Dense activation are not necessarily desirable, so we also inspect the marginal utilization of each MLP activation dimension: how often a dimension is activated across tokens (thresholded at 0.1).
In the mechanistic interpretability literature, low marginal utilization and ``dead'' dimensions are strong indicators of under-utilized capacity~\citep{Brickenetal2023,gao2025scaling}.
Figure~\ref{fig:activation-utilization} shows this quantity across layers. While neither model has a perfectly even distribution, \modelname consistently achieves substantially better load-balancing between activation dimensions. In \S\ref{sec:additional-results}, we also show a similar trend if we additionally take activation magnitude into account, demonstrating that \modelname more evenly uses the activation dimensions in the middle layers of the network.

\subsection{\modelname Mitigates Middle-layer Representation Collapse} \label{sec:analysis-residual}

We now turn from studying MLP activations to the residual stream after each layer.
Recent analyses of deep, constant-width LMs reveal the emergence of ``compression valleys,'' where the LM's middle layers collapse in representational capacity, characterized by a severe drop in representational entropy~\citep{skean2025layer,queipo-de-llano2026attention}. Following \citet{queipo-de-llano2026attention}, we track the normalized matrix entropy of the residual stream across all layers:
\begin{equation}
    \frac{1}{\log r} \left( -\sum_{j=1}^{r} p_j \log p_j\right), \quad p_j = \sigma_j^2 / \|\mathbf{X}\|_F^2
\end{equation}
where $\sigma_j$ are the sorted singular values of the input-feature representation matrix $\mathbf{X}$ with rank $r$, again computed using the WikiText-2 validation split.
Closely related to the effective dimension metric~\citep{hill,7098875},
a higher matrix entropy indicates a more ``even'' use of the representation space.

We consider the per-layer hidden states in this analysis. For \modelname, recall our interpretation from \S\ref{sec:method} that considers it having a wide residual stream where each layer reads from/writes to only a subset of dimensions. Accordingly, we consider this wide residual stream as its effective hidden states.

In Figure~\ref{fig:matrix-entropy}, we see that the baseline model exhibits a severe compression valley: in middle layers, its normalized entropy drops to near-zero, indicating that the token representations have collapsed into a highly degenerate, low-rank subspace despite the large width.
This is consistent with prior findings ~\citep{skean2025layer,queipo-de-llano2026attention}.
In contrast, \modelname restructures this dynamic. While it actively lowers its entropy in the early layers to compress the representation (anticipating the width reduction), it avoids the middle-layer collapse. Throughout the bottleneck and final layers, \modelname maintains a higher normalized entropy, potentially suggesting that physically constraining the parameter space encourages the network to maintain a high-entropy manifold.

\subsection{Predictive Dynamics via the Logit Lens}
\label{sec:analysis-logit-lens}

\begin{figure}[t]
\vspace{-4mm}
    \centering
    \includegraphics[width=\linewidth]{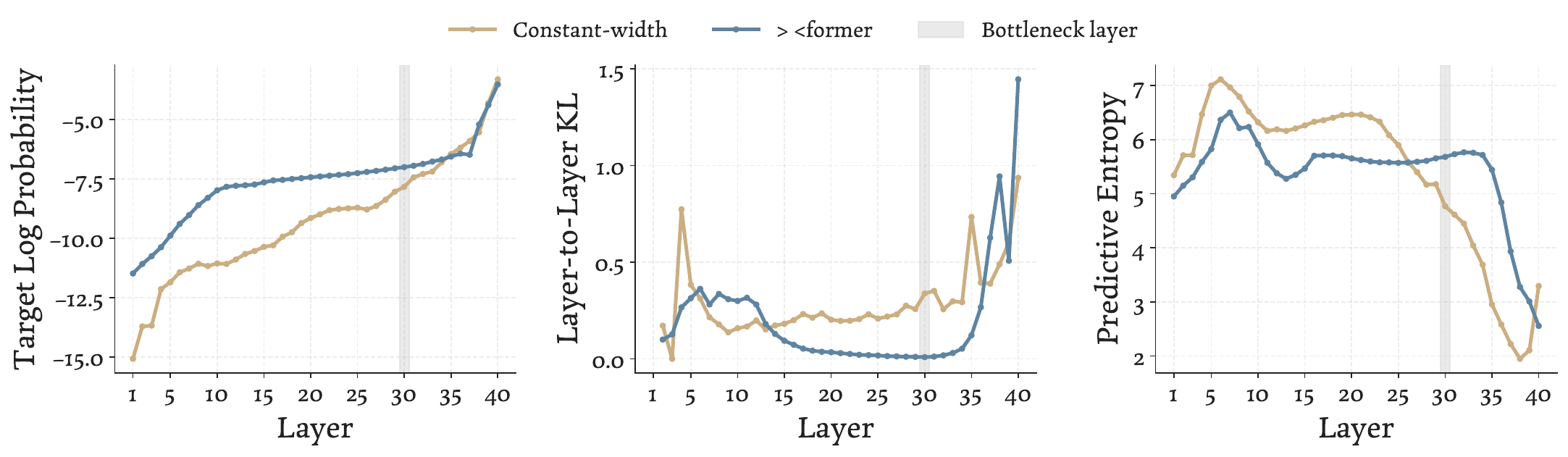}
    \caption{Logit lens analysis of the 2B \modelname versus the constant-width baseline. \textbf{Left:} \modelname assigns higher target-token probability through much of the network. \textbf{Middle:} \modelname's decoded token distribution changes more gradually across middle layers. \textbf{Right:} \modelname has lower entropy in early layers but declines less rapidly than the baseline in the final layers.}
    \label{fig:logit-lens}
    \vspace{-2mm}
\end{figure}

To understand how these geometric differences in the residual stream affect model predictions, we project intermediate hidden states into vocabulary space using the logit lens~\citep{logit-lens}. Specifically, at each layer, we decode hidden states by applying the final RMS normalization followed by the unembedding matrix. As in \S\ref{sec:analysis-residual}, we treat \modelname's effective wide residual stream as its hidden state, restricting to the residual dimensions visible to the unembedding.

For each layer, we measure the target-token log probability, the entropy of the decoded token distribution, and the layer-to-layer KL divergence between adjacent logit-lens distributions. We symmetrize this KL by averaging the two directions, using it as a proxy for how rapidly the decoded distribution changes with depth.

Figure~\ref{fig:logit-lens} shows that \modelname assigns higher probability to the target token, with lower decoded-distribution entropy, through much of the early-to-middle network. At the same time, its decoded token distribution changes more gradually across layers, as reflected by lower layer-to-layer KL. In the final layers, the distribution changes rapidly again as probability mass concentrates on the target token.

\begin{wraptable}[11]{R}{4.75cm} %
    \centering
       \vspace{-4mm}
    \begin{tabular}{lr} %
        \toprule
        \textbf{Expansion Method} & \textbf{Loss} \\
        \midrule
        Constant-width    & 3.138 \\
        \midrule
        Carry-forward     & 3.099 \\
        Zero Padding        & 3.124 \\
        Projection & 3.150 \\
        \bottomrule
    \end{tabular}
    \caption{Performance comparison of different methods to expand extra dimensions at 500M. \textbf{Simply carrying forward features from lower layers performs the best.}}
    \label{tab:moveup-ablation}
        \vspace{-4mm}
\end{wraptable}

\subsection{Ablations} \label{sec:moveup-ablation}

We analyze alternative methods of expanding dimensions.
Beyond our default method that carry forward features by copying coordinates through the residual stream, we also consider (1) padding with 0s and (2) training a projection layer to predict the extra dimensions from the previous layer representation.\footnote{We also tried training a projection to predict the entire new layer representation, not just the \emph{extra} dimensions, but it is empirically unstable and diverges during training.}
For each, we also sweep over multiple hyperparameter configurations and report the best loss.
We ablate at the 500M scale, and Table~\ref{tab:moveup-ablation} shows that copying features performs the best.

\section{Limitations}

A major caveat is that our approach adds significant complexity for efficient training.
Concretely, for efficient training one would need to develop and optimize kernels for many different shapes, each of which has different latency, memory footprint, and compute profiles. The fixed-residual construction also potentially adds overhead, since the slicing, copying, and zero-padding around a global residual stream wider than the baseline $d$ introduce extra kernel launches, though much of this could be mitigated via kernel fusion. Heterogeneous per-layer widths are further in tension with standard tensor/pipeline parallelism techniques.

We stress, however, that these are
implementation rather than algorithmic limitations: variable-width transformers are still matmul-rich, and the gap we describe reflects the fact that
current infrastructure has been heavily optimized for the uniform-width regime rather than any intrinsic property of the architecture. We expect that purpose-built kernels would close much of the gap between theoretical and realized efficiency. 

More broadly, while we are not calling for immediate adoption of \modelname{s}, we hope that future architecture research can capitalize on this previously unnoticed degree of freedom in design.

\section{Related Work}

\paragraph{Nonuniform allocation of width in transformers.}
Several transformer variants allocate parameters nonuniformly across depth. DeLighT uses block-wise scaling, making earlier blocks shallower/narrower and later blocks deeper/wider \citep{mehta2020delight}. OpenELM adopts layerwise scaling in decoder-only language models by varying attention and feed-forward dimensions across layers \citep{mehta2024openelm}. Recent layerwise-scaling variants explore framed, reverse, and crown allocation profiles \citep{baroian2025crown}. \citet{ikeda2025layerwise} study the layerwise importance of feed-forward networks by reallocating MLP capacity and find benefits from concentrating MLPs in middle layers. Our work differs from these approaches in varying the full block hidden dimension, rather than only the attention-head count, MLP multiplier, or lightweight block internals. This requires addressing how variable-width blocks interact with the residual stream; our fixed-residual construction lets inactive coordinates bypass narrower blocks.

\paragraph{Bottleneck across sequence length.}
There has also been work that performs compression across sequence length. Funnel-Transformer gradually shortens the sequence of hidden states and later recovers token-level representations for prediction \citep{dai2020funnel}. Hourglass Transformers downsample and upsample activations to build an explicit hierarchical language model \citep{nawrot2022hierarchical}. Perceiver models use cross-attention to distill high-dimensional inputs into a compact latent bottleneck before applying transformer-style processing \citep{jaegle2021perceiver}. These methods primarily bottleneck the number of tokens or latent slots. Our architecture instead preserves the token sequence length and introduces a bottleneck in hidden width across depth.

\paragraph{Bottleneck designs outside Transformers.}
Bottleneck architectures have a long history outside of transformers. The U-Net and stacked hourglass networks use encoder--decoder structures that repeatedly reduce and recover spatial resolution, often with skip connections that preserve high-resolution information \citep{ronneberger2015u,newell2016stacked}. Other architectures introduce bottlenecks along the channel dimension: ResNets use bottleneck residual blocks to reduce the cost of deep convolutional networks \citep{he2016deep}, while MobileNetV2 uses inverted residual blocks with linear bottlenecks for efficient vision models \citep{sandler2018mobilenetv2}. However, transformer applications have mostly worked with non-bottleneck architectures in the channel dimension.

\paragraph{Hyper-Connections.}

By expanding residual-stream capacity, \modelname is conceptually related to Hyper-Connections (HC) \citep{zhu2025hyperconnections,xie2026mhcmanifoldconstrainedhyperconnections,deepseekai2026deepseekv4}. However, the mechanisms are different: HC uses learned mixing between multiple residual streams, whereas \modelname uses deterministic slicing and carry-forward within a single global residual stream. In narrower layers, inactive coordinates bypass the block and are reintroduced when the width expands. Thus, \modelname provides a complementary way to vary residual capacity without the learned residual-mixing matrices that \citet{xie2026mhcmanifoldconstrainedhyperconnections} identify as a source of large-scale HC instability.

\section{Conclusion}

In this work, we challenge the standard assumption of uniform capacity allocation across transformer depth by introducing the \modelname, a variable-width architecture. Across evaluations from 200M to 3B parameters (dense and MoE), parameter-matched \modelname{s} outperform uniform baselines, while mathematically and empirically reducing both FLOPs and KV cache memory. Furthermore, our analyses reveal that this bottleneck design may act as a structural regularizer, forcing the network to utilize its representation space more evenly. These findings demonstrate that nonuniform width allocation is an efficient and promising strategy for scaling future language models.

\section*{Acknowledgments}
This study was supported in part by the MIT-IBM Watson AI Lab and the National Science Foundation under CAREER Award No. 2441872 and NSF grant No. CCF-21-12665.

\bibliography{references}
\bibliographystyle{bib-style}

\appendix

\section{Parameter-Matched Width Calculation}
\label{app:param-match}

Here we derive how we instantiate a geometric width schedule from a
bottleneck layer $\ell^*$ and bottleneck dimension $d_{\ell^*}$, while
matching the parameter count of a constant-width baseline. The derivation is in
continuous width space; integer rounding is applied only after the
parameter-matched widths have been determined.

Let $L$ be the number of transformer layers, $d$ the hidden dimension of the
constant-width baseline, and $v$ the vocabulary size. We assume the input
and output embeddings maintain the baseline width $d$. The residual
stream has a layer-dependent width $d_\ell$, and resizing between adjacent
layers is parameter-free. We impose symmetric endpoint widths, $d_1=d_L=\bar{d}$.

The layer widths follow a geometric progression on each side of the bottleneck:
$$
    d_\ell =
    \begin{cases}
        \alpha^- d_{\ell-1}, & 1 < \ell \leq \ell^*,\\
        \alpha^+ d_{\ell-1}, & \ell^* < \ell \leq L,
    \end{cases}
$$
where $\alpha^- \leq 1$ and $\alpha^+ \geq 1$. Symmetric endpoints imply $(\alpha^-)^{\ell^*-1}(\alpha^+)^{L-\ell^*}=1$, which constrains $\alpha^+$ for any candidate $\alpha^-\in(0,1]$:
$$
    \alpha^+ = (\alpha^-)^{-\frac{\ell^*-1}{L-\ell^*}}.
$$
Thus, the shape is strictly determined by $\alpha^-$. We define the dimensionless factors $c_\ell(\alpha^-)$ such that $d_\ell = \bar{d}\,c_\ell(\alpha^-)$:
$$
    c_\ell(\alpha^-) =
    \begin{cases}
        (\alpha^-)^{\ell-1}, & 1 \leq \ell \leq \ell^*,\\
        (\alpha^-)^{\ell^*-1}(\alpha^+)^{\ell-\ell^*}, & \ell^* < \ell \leq L.
    \end{cases}
$$

To match the parameter count of the constant-width baseline, we must account for the dominant layer parameters and endpoint corrections. For a dense transformer block with SwiGLU, the per-layer parameter count scales with $K d_\ell^2$, where $K = 4 + N_m E$ ($E=4$ is the MLP expansion factor, and $N_m=3$ for SwiGLU is the number of MLP projections). Ignoring layer norm and bias terms, the baseline parameter count is $P_{\mathrm{base}} = 2vd + LKd^2$.

Because the embeddings are fixed at width $d$, if our schedule requires $\bar{d} > d$ (as is the case for \modelname), we pad the initial embeddings with 0s and truncate the final unembeddings. This results in unused parameters in the first attention layer and the final MLP layer. Specifically, the first attention QKV map and the last MLP output map contain $3\bar{d}(\bar{d}-d)$ and $E\bar{d}(\bar{d}-d)$ unused parameters, respectively. The total endpoint correction is therefore:
$$
    W_{\mathrm{end}}(\bar{d}) = \mathbf{1}\{\bar{d}>d\}\,(3+E)\bar{d}(\bar{d}-d).
$$

Equating the valid parameters of our variable-width model to the baseline gives:
$$
    K\bar{d}^2 S_2(\alpha^-) - W_{\mathrm{end}}(\bar{d}) = LKd^2,
$$
where $S_2(\alpha^-) = \sum_{\ell=1}^{L} c_\ell(\alpha^-)^2$.

Substituting in $W_{\mathrm{end}}(\bar{d})$, we can simplify this to:
$$
    \Big[ K S_2(\alpha^-) - \mathbf{1}\{\bar{d}>d\}\,(3+E) \Big] \bar{d}^2 + \Big[ \mathbf{1}\{\bar{d}>d\}\,d(3+E) \Big] \bar{d} - LKd^2 = 0.
$$
Because the coefficients depend on the piecewise indicator $\mathbf{1}\{\bar{d}>d\}$, we solve this by assuming a state for the indicator (either $0$ or $1$), applying the standard quadratic formula to find the positive root, and selecting the root that is self-consistent with our assumption. This expresses the valid endpoint width as a function of $\alpha^-$, which we denote as $\bar{d}_{\alpha^-}$. The bottleneck width for the given $\alpha^-$ is then:
$$
    b(\alpha^-) = \bar{d}_{\alpha^-}(\alpha^-)^{\ell^*-1}.
$$

We use a 1D numerical solver over $\alpha^- \in (0,1]$ to solve for $b(\alpha^-)=d_{\ell^*}$, where $d_{\ell^*}$ is the desired bottleneck dimension. Finally, the continuous widths are rounded to the nearest multiple of the attention head dimension $Q$ to ensure compatibility:
$$
    \widehat{d}_\ell = \mathrm{round}\!\left(\frac{d_\ell}{Q}\right)Q.
$$

\section{Dataset Statistics} \label{sec:dataset-stats}

In this section, we report the statistics of the datasets we used in our evaluation in Table~\ref{tab:dense-moe-eval}.

\begin{table}[h!]
\centering
\caption{Task evaluation configurations and metrics.}
\begin{tabular}{llrl}
\toprule
\textbf{Task} & \textbf{Domain} & \textbf{Split / Instances} & \textbf{Metric} \\
\midrule
OpenBookQA & science QA & 500 & \texttt{acc\_norm} \\
PIQA & physical commonsense & 1,838 & \texttt{acc\_norm} \\
SciQ & science QA & 1,000 & \texttt{acc\_norm} \\
ARC-Easy & grade-school science QA & 2,376 & \texttt{acc\_norm} \\
ARC-Challenge & difficult science QA & 1,172 & \texttt{acc\_norm} \\
BoolQ & yes/no reading comprehension & 3,270 & \texttt{acc} \\
COPA & causal commonsense & 100 & \texttt{acc} \\
HellaSwag & commonsense completion & 10,042 & \texttt{acc\_norm} \\
WinoGrande & pronoun/coreference reasoning & 1,267 & \texttt{acc} \\
RACE & reading comprehension & 1,045 & \texttt{acc} \\
WikiText & language modeling & 62 documents & perplexity \\
LAMBADA OpenAI & long-context word prediction & 5,153 & \texttt{acc}, perplexity \\
\bottomrule
\end{tabular}
\label{tab:downstream-eval-stats}
\end{table}

\section{Additional Results} \label{sec:additional-results}

\begin{wrapfigure}[22]{R}{6.2cm} %
    \centering
    \includegraphics[width=\linewidth]{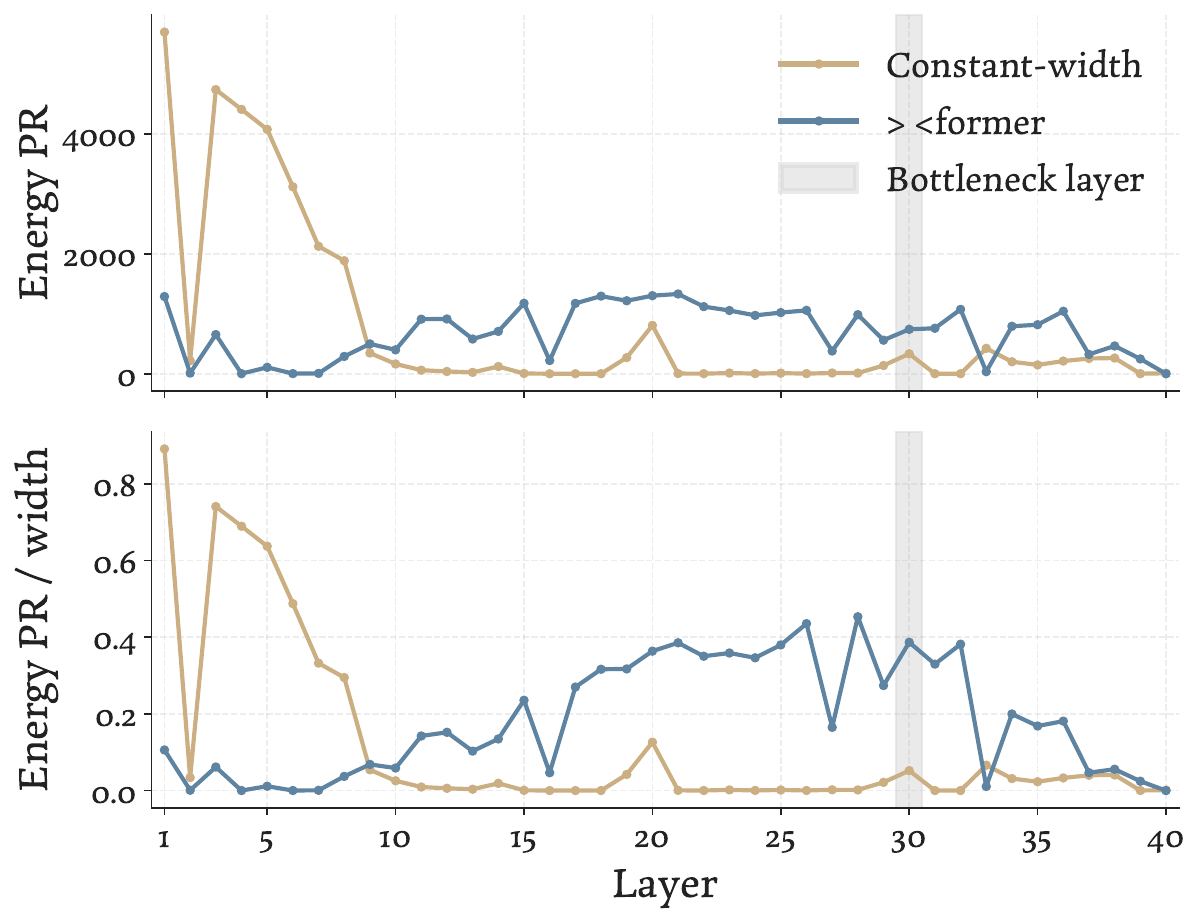}
    \caption{The Participation Ratio (PR; \S\ref{sec:analysis-activation}) of MLP activations in the 2B \modelname vs. the 2B constant-width transformer. We show both the raw PR and the normalized PR by the layer width. \textbf{\modelname has a higher PR in the middle layers, corresponding to more even usage of the activation dimensions in those layers.}}
    \label{fig:activation-PR}
\end{wrapfigure}

The analysis in \S\ref{sec:analysis-activation} shows that \modelname achieves better activation density, but it does not account for activation magnitude. Large language models frequently develop severe outlier dimensions, rendering the remaining active dimensions computationally insignificant. To evaluate this, we compute the energy Participation Ratio (PR) over the MLP activations~\citep{LITWINKUMAR20171153,2jt7-c8cq}. Let $a_{t,i}$ denote the activation of dimension $i$ for token $t$ and $e_i = \sum_t a_{t,i}^2$ denote the total energy of dimension $i$ across all tokens $t$. The effective number of utilized dimensions is given by $N_{\text{eff}} = (\sum_i e_i)^2 / \sum_i e_i^2$.
Intuitively, $N_{\text{eff}}$ acts as a continuous measure of representational equality: if a single outlier dimension hoards all the numerical energy, $N_{\text{eff}}$ collapses to 1, regardless of the actual width $d_\ell$. Conversely, if computational energy is distributed uniformly across all coordinates, $N_{\text{eff}}$ reaches its theoretical maximum of $d_\ell$. Thus, computing the width-normalized fraction ($N_{\text{eff}}/d_\ell$) provides a measure of the fraction of effectively utilized dimensions. We report both the absolute and normalized PR in Figure~\ref{fig:activation-PR}. 

The results reveal a distinction between \modelname and the constant-width transformer. For the baseline, its width-normalized energy utilization collapses to near zero ($<5\%$) by around layer 10. In contrast, \modelname more evenly distributes energy utilization in the middle layers by maintaining an absolute PR of roughly 1,000 effective dimensions, yielding a richer representation manifold. By restricting parameter availability, the bottleneck in \modelname may act as a structural regularizer that encourages the network to pack a denser representation into the available capacity.

\end{document}